%% file: 00_main.tex
\documentclass[10pt,twocolumn,letterpaper]{article}

\usepackage{cvpr}
\usepackage{times}
\usepackage{epsfig}
\usepackage{graphicx}
\usepackage{amsmath}
\usepackage{amssymb}
\usepackage{bbm}
\usepackage{enumitem, url}
\usepackage{comment}
\usepackage{subcaption}
\usepackage{lipsum}
\usepackage{textcomp}
\usepackage{color}
\usepackage{graphicx,calc}
\usepackage{listings}
\usepackage{anyfontsize}
\usepackage[title]{appendix}

\newcommand{\modellong}{Hierarchical Planning and Reinforcement Learning}
\newcommand{\model}{HIP-RL}

\newlength\myheight
\newlength\mydepth
\settototalheight\myheight{Xygp}
\newcommand*\inlinegraphics[1]{%
  \settototalheight\myheight{Xygp}%
  \settodepth\mydepth{Xygp}%
  \raisebox{-\mydepth}{\includegraphics[height=\myheight]{#1}}%
}

\usepackage[pagebackref=true,breaklinks=true,letterpaper=true,colorlinks,bookmarks=false]{hyperref}

\cvprfinalcopy 

\def\cvprPaperID{6137} 

\ifcvprfinal\fi
\begin{document}

\nocite{thomason2018shifting}

\title{What Should I Do Now?\\Marrying Reinforcement Learning and Symbolic Planning}

\author{Daniel Gordon$^1$ \quad Dieter Fox$^{1,2}$ \quad Ali Farhadi$^{1,3}$\\
{\normalsize $^1$Paul G. Allen School of Computer Science, University of Washington} \\
{\normalsize $^2$Nvidia} \ 
{\normalsize $^3$Allen Institute for Artificial Intelligence}
}

\maketitle

\begin{abstract}
Long-term planning poses a major difficulty to many reinforcement learning algorithms. This problem becomes even more pronounced in dynamic visual environments. In this work we propose \modellong{} (\model{}), a method for merging the benefits and capabilities of Symbolic Planning with the learning abilities of Deep Reinforcement Learning. We apply \model{} to the complex visual tasks of interactive question answering and visual semantic planning and achieve state-of-the-art results on three challenging datasets all while taking fewer steps at test time and training in fewer iterations. Sample results can be found at \url{youtu.be/0TtWJ_0mPfI}\footnote{The full dataset and code will be open sourced soon.}
\end{abstract}

\input{01_intro}

\input{02_relatedwork}
\input{03_method}
\input{04_tasks}

\input{05_experiments}
\input{06_conclusion}


{\small
\bibliographystyle{ieee}
\bibliography{references}
}

\input{07_appendix}

\end{document}

%% file: 01_intro.tex
\vspace{-6mm}
\section{Introduction}

An important goal in developing systems with visual understanding is to create agents that interact intelligently with the world. Teaching these agents about the world requires several steps.
An agent must initially learn simple behaviors such as navigation and object affordances. Then, it can combine several actions together to accomplish longer term goals. As the task complexity increases, the agent must plan farther in the future.

\begin{figure}[t]
\begin{center}
\includegraphics[width=1.0\linewidth]{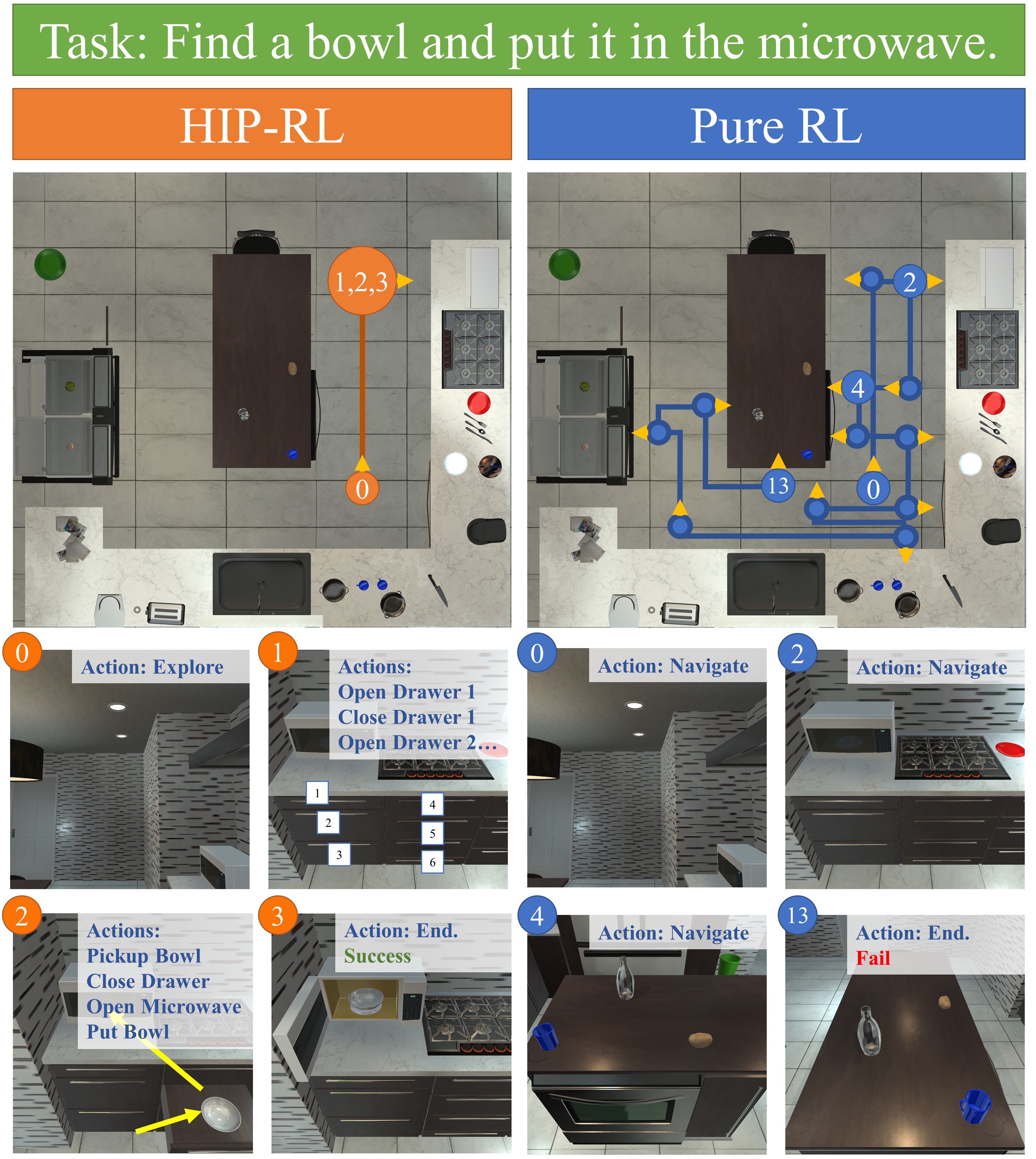}
\vspace{-7mm}
\caption{Sample Visual Semantic Planning task execution. The agent is asked to put a bowl in the microwave. At t=0, \model{} has not observed any locations where bowls can be, so it explores the room. At t=1, the Meta-Controller invokes the Planner which creates an efficient plan to check all the cabinets. It also sees the microwave and saves this information for later. At t=2 a bowl is found, and the Planner updates its plan to finish the task. Since \model{} already saw the microwave, it saves time by not needing an additional search. At t=3 the Planner puts the bowl in the microwave, returning control to the Meta-Controller which finishes the episode. In contrast, the Pure RL system spends much more time exploring the room, does not open any drawers, and ends the episode after many more steps, failing the task.} 
\label{fig:teaser}
\end{center}
\vspace{-7mm}
\end{figure}

In recent years, researchers have predominantly trained interactive agents using either deep learning techniques on raw visual data or using planning algorithms on symbolic state representations. Deep learning has proven to be a very useful tool at learning to extract meaningful features from large sources of raw data. Deep Reinforcement Learning (Deep RL) has gained significant traction in the vision community for simple tasks like playing video games~\cite{a3c, curiosity}. Yet as task complexity increases, and longer-term planning is required, these systems can no longer learn good reactive policies due to the exponentially branching state space.

Conversely, many robotics systems still favor planning and search techniques such as RRTs and A* over the reactive Deep RL counterparts~\cite{ chitta2012moveit, sucan2012open}. The traditional planning algorithms offer better generalization when sensor data is clean, and many provide optimality guarantees which are beneficial for ensuring safety in potentially dangerous robotics environments. However most planning algorithms assume either perfect or fairly accurate state estimation and cannot operate on high dimensional raw sensor input. This is especially true for task planning algorithms which use binary values as state representations (e.g. the apple \textit{is} in the fridge) like Planning Domain Definition Language (PDDL) solvers.

As such, deep learning and task planning have complementary benefits and drawbacks: deep learning techniques can extract information from raw (pixel) data but fail at long-term planning, and task planning require preprocessed data but can use it to construct multi-step plans which satisfy complex goals. 

Much of the recent work in this area has treated learning and planning as separate problems. We present \modellong{} (\model{}), a method for merging these techniques to benefit from the strengths of both. \model{} has several advantages over traditional planning as well as over current Deep RL techniques. 1) Due to the correctness/completeness guarantees of the planner, \model{} increases the accuracy and effectiveness over comparable pure RL systems. 2) By relying on a planner to create sequences of actions, \model{} significantly reduces the number of steps that an agent takes at test time. 3) By simplifying the learning procedure and shortening the path lengths, we are able to train our algorithm using an order of magnitude fewer training examples. 4) \model{} can also learn to account for noisy sensor data which may otherwise hinder a symbolic planner. 

To evaluate the usefulness and effectiveness of our algorithm, we apply \model{} to a variety of tasks. First, we apply \model{} to the task of Interactive Question Answering (IQA), an extension of Visual Question Answering (VQA) where an agent dynamically navigates in and interacts with an environment in order to answer questions. We apply \model{} to IQUAD~V1~\cite{iqa} and EQA~V1~\cite{eqa} and achieve state-of-the-art results on both tasks. We additionally show that \model{} is able to perform complex Visual Semantic Planning (VSP) tasks such as \texttt{Find a bowl and put it in the microwave}, dramatically outperforming both learning-only and planning-only baselines. In general, we find that using planning and learning together results in higher accuracy, more efficient exploration, and faster training than other methods.

%% file: 02_relatedwork.tex
\section{Related Work}

\subsection{Planning and Learning}
\vspace{-2mm}
Although both learning and planning have significant amounts of prior work, there have been relatively few attempts at merging the two. Many recent reinforcement-learning based algorithms fail when long-term planning is required; most algorithms trained on ATARI fail on the Montezuma's Revenge game due to its sparse rewards and long episodes~\cite{a3c}. Yet when planning and learning are combined, the results are often greater than either could do alone. One example of successfully merging planning and learning is the AlphaZero family of algorithms which combine Deep RL for board state evaluation and Monte Carlo Tree Search (MCTS) for planning and finding high-value future board positions~\cite{alphazero}. Rather than using MCTS to intelligently explore future states, which is not feasible in a partially observed visually dynamic environment, we use the Metric-FF Planner~\cite{metricff} to plan a single trajectory to the goal state. This chains actions together in order to shorten the number of hierarchical decisions and reduce the size of the action space. Planning in stochastic environments is often solved using Partially Observable Markov Decision Processes. Although they are frequently used to great success~\cite{amato2016policy, cai11hyp}, POMDPs often assume a known noise and transition model, which is not readily applicable for algorithms which use deep feature extraction. We avoid this issue by using both planning and learning; although our planner operates under the assumption of perfect and complete information, the Meta-Controller can divert control to the learning-based methods in the event of a planner failure or to gather more information.

\subsection{Hierarchical Reinforcement Learning (HRL)}
\vspace{-2mm}
HRL seeks to solve several problems with standard reinforcement learning such as handling very long episodes with sparse rewards. The design of these systems typically has one hierarchical meta-controller which invokes one of several low-level controllers. Each low-level controller is trained to accomplish a simpler task. In many cases~\cite{ eqa, das2018neural, iqa, kulkarni2016hierarchical, tessler2017deep} both the meta-controller and all low-level controllers are learned, and in some cases~\cite{kulkarni2016hierarchical, tessler2017deep} the tasks of the sub-controllers are also learned purely from interactions during training episodes rather than being human-engineered. This allows these systems to generalize well to unseen tasks with only a few training examples for the new tasks. In contrast, we use some learned low-level controllers, and some which use planning algorithms to directly solve subtasks. This allows our system to train qickly and still generalize well to new tasks with only moderate additional goal-state specification.

\subsection{Deep RL in Virtual Visual Environments}
\vspace{-2mm}
In the past few years, many different virtual platforms have been created in order to facilitate better Deep RL. Virtual environments provide limitless labeled data and are easily parallelizable. Some of the most popular are OpenAI Gym~\cite{openaigym} and VizDoom~\cite{kempka2016vizdoom} which both build on existing video games, and MuJoCo~\cite{mujoco} which implements more realistic contact physics. More recently, multiple environments have been created which offer near-photo-realistic and physically accurate interaction such as AI2-THOR~\cite{ai2thor}, Gibson~\cite{gibson}, and CHALET~\cite{yan2018chalet}. Other environments forgo photo-realism for increased rendering speed such as House3D~\cite{house3d}, and DeepMind Lab~\cite{deepmindlab}.

Additionally, there have been many advancements in learning from interactions with a virtual visual environment. Recent works have used virtual environments for Question Answering~\cite{eqa, das2018neural, iqa}, visually-driven navigation~\cite{gupta2017cognitive, Mirowski2016LearningTN, zhu2017target}, and semantic navigation~\cite{anderson:cvpr18, Chaplot2018GatedAttentionAF}. However, much of the focus has been on improving navigation techniques in these environments. We are interested in expanding beyond navigation to more complex visual tasks which require both navigation and long term planning. In this work, we use the existing IQUAD~V1\footnote{https://github.com/danielgordon10/thor-iqa-cvpr-2018} dataset presented in~\cite{iqa} which is built on the AI2-THOR~\cite{ai2thor} environment, and EQA~V1\footnote{https://github.com/facebookresearch/EmbodiedQA}~\cite{eqa} which uses the House3D environment~\cite{house3d}. Additionally, we construct a new dataset for Visual Semantic Planning built on AI2-THOR, explained more in section \ref{section:vsp}.

\subsection{Task and Motion Planning (TaMP)}
\vspace{-2mm}
Task and motion planning is the problem of accomplishing goals at a high level by planning the exact motion trajectories for a robot. Much of the work in TaMP uses hierarchical algorithms to plan high level actions for long-term goals and use motion planning algorithms for fine-grained motor manipulation. \cite{kaelbling2011hierarchical} outlines a detailed hierarchy for planning and executing simple manipulation tasks. \cite{konidaris2018skills}  attempt to learn symbolic representations by interacting with an environment and observing the effects of actions. \cite{chitnis2016guided} learn heuristics to shorten the search procedure for their TaMP algorithm. We differ from these approaches in that we assume perfect manipulators, which simplifies our control setup, but use pixel-level inputs from near-photo-realistic 3D simulation environments. We also do not assume complete state information is given to the controller. Finally, we use reinforcement learning to direct our hierarchical controller, whereas most methods treat the entire state as fully observable and therefore plan and execute complete motion trajectories from the initial state.

%% file: 03_method.tex
\section{Method}
In order to accomplish complex tasks, a system must be able to plan long action trajectories which satisfy the task goals. To operate in a visual world, a system must learn to understand a dynamic visual environment. Thus, to learn to plan in a visual environment, we combine Deep RL with Symbolic Planning, handing control of the agent back and forth between the two methods. In this section we outline the individual components of \model{} as well as how they work together to solve complex learning and planning tasks.

\begin{figure}[t]
\begin{center}
\includegraphics[width=1.0\linewidth]{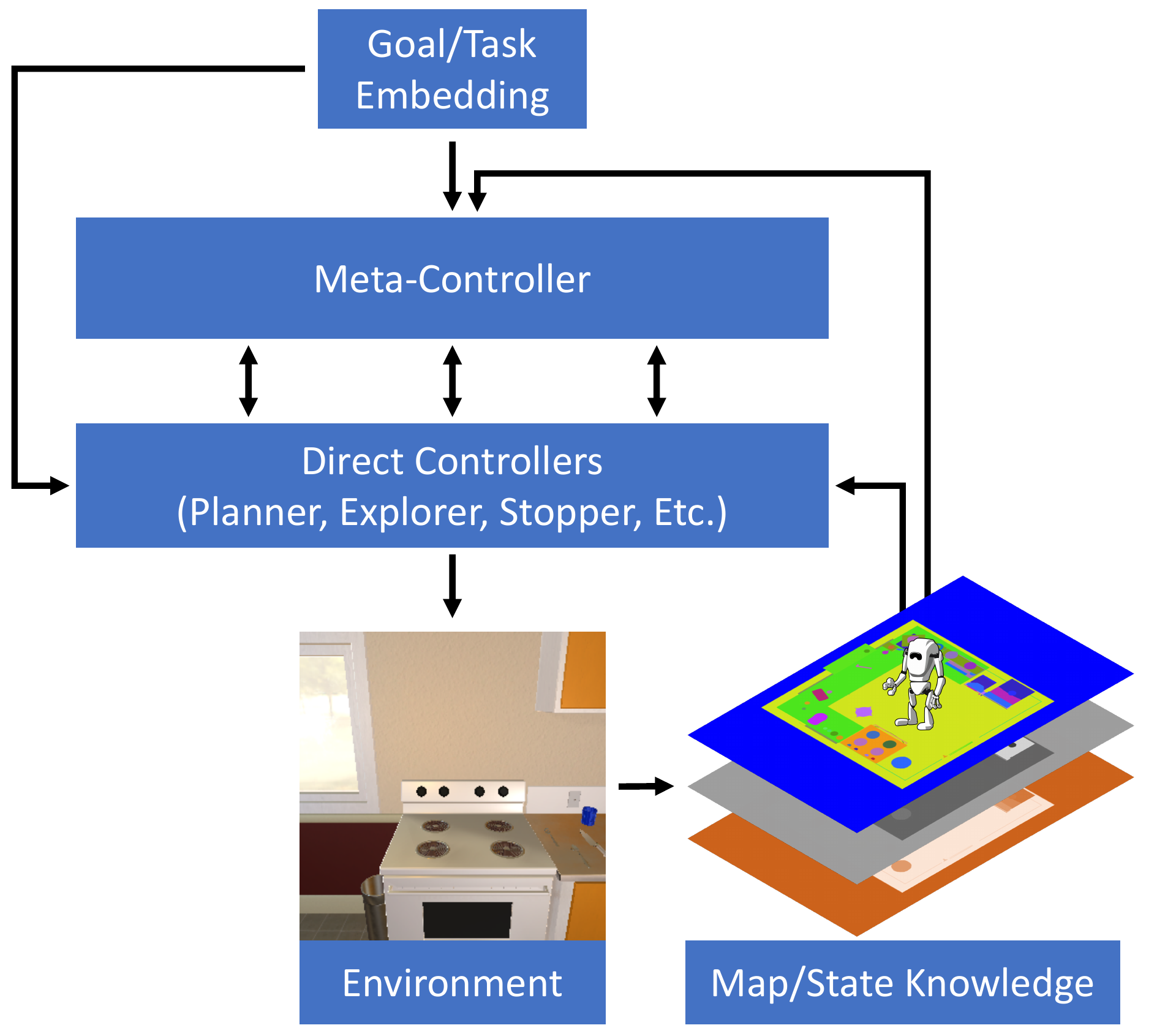}
\vspace{-3mm}
\caption{Diagram of the \model{} framework. Each direct controller interacts with the environment based off of commands given by the Meta-Controller. All controllers share a knowledge state which is updated by the various controllers during the episode based on perceptual and interactive observations.} 
\label{fig:hierarchy}
\end{center}
\vspace{-8mm}
\end{figure}

\subsection{\modellong{} (\model{})}
\vspace{-2mm}
\model{} consists of a hierarchical Meta-Controller, several direct (low-level) controllers, and a shared knowledge state (Figure \ref{fig:hierarchy}). The knowledge state contains all perceptual and interactive information such as navigable locations, object positions, and which cabinets have previously been opened, as well as the goal representation. For example, in Figure \ref{fig:teaser} the knowledge state in image \inlinegraphics{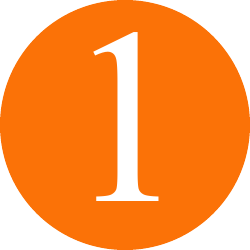} contains the positions of the drawers, the microwave, the red plate, and the fact that none of the drawers have been checked. Each controller can read all of the state knowledge, and can update a portion of the knowledge based on its perception; the Navigator can update the world map, and the Object Detector can update the object locations but not vice versa. At the start of an episode, the Meta-Controller chooses which direct controller should be used to advance the current state towards the goal state, invoking that direct controller with a subtask. The direct controller attempts to accomplish this subtask and returns control back to the Meta-Controller upon completion. This process is repeated until the Meta-Controller decides the full task has been accomplished and calls the Stopper to end the episode. 

\begin{figure*}[t]
\begin{center}
\includegraphics[width=0.9\linewidth]{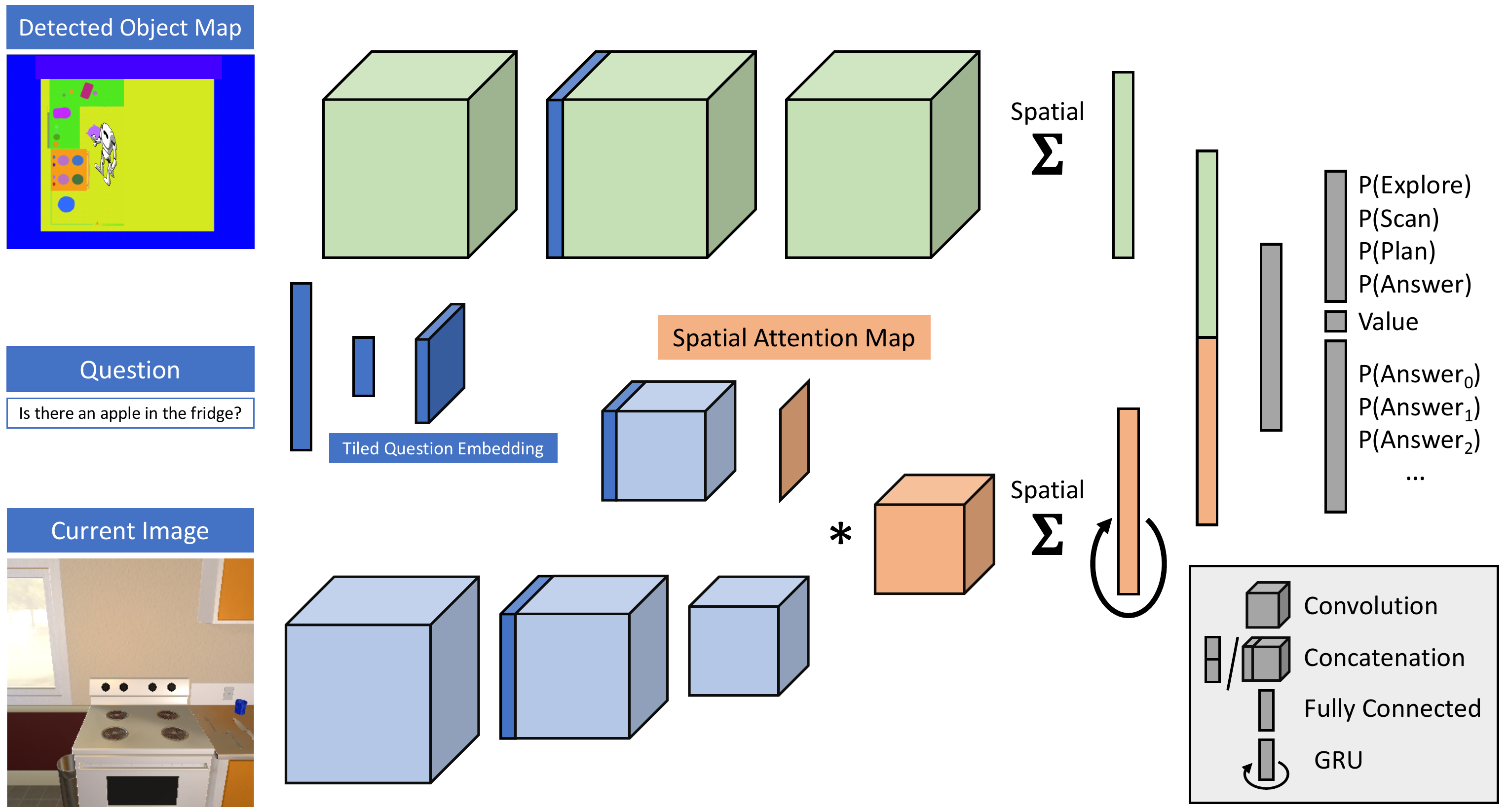}
\vspace{-3mm}
\caption{Overview of the network architecture for the hierarchical Meta-Controller (and answerer) used for IQA tasks. The network takes as input the full detected object map, the question, and the current image. The question is embedded and spatially tiled. We concatenate the object map features with the question embedding, perform several convolutions, and spatially sum the output. Similarly, we concatenate the image features with the question embedding, and use an attention mechanism conditioned on the question to spatially sum the features. We do not use an attention mechanism on the detected object map as this makes counting questions difficult. The image features additionally use a GRU~\cite{cho2014learning} to add temporal context. Since the detected object map is purely additive (the final map contains at least as much information as the previous ones) no temporal context is necessary. The map and image features are concatenated and fed into a fully-connected layer. The network outputs a probability distribution over the actions, a value estimate for the state, and a distribution over the answers for the question.} 
\label{fig:network}
\end{center}
\vspace{-8mm}
\end{figure*}

\subsection{Hierarchical Meta-Controller}
\vspace{-2mm}
The Meta-Controller receives the goal and decides which of the direct controllers to invoke. It learns to trade off between the length of an episode and the reward/penalty received for successfully or unsuccessfully ending an episode. We train this behavior using the A3C algorithm \cite{a3c} to reward successful episodes, penalize unsuccessful episodes, and add a time penalty for each hierarchical step. We visualize the Meta-Controller's architecture in the context of Interactive Question Answering in Figure \ref{fig:network}. It takes as input a top-down spatial map of object locations, the question representation, and the current image from the environment. The question embedding is concatenated with the spatial map and visual features to condition the features on the question. The network outputs a probability distribution over the action space of direct controllers as well as a value for the current state. The Meta-Controller architecture for Visual Semantic Planning is exactly the same except the question embedding is replaced with the semantic task embedding, and there is no answer branch in the output.

\subsection{Planner}
\vspace{-2mm}
The Planner is tasked with returning a sequence of actions which would accomplish the goal. It operates on logical primitives using the Planning Domain Definition Language (PDDL) and is guaranteed to return a correct set of high-level planning actions given the observations are accurate. PDDL specifies states using ``fluents'' which can be boolean values (cabinet 1 is open, an apple is in the fridge) or numeric (location A and B are 10 steps apart). Actions consist of templatized preconditions necessary for the action to be possible, and effects, which modify the values of the fluents caused by executing that action. For example, the \texttt{Open} action takes the preconditions that an object must be openable but closed and that the agent must be near the object, and has the effect of setting the object's \texttt{closed} fluent to \texttt{False}. Goals specify a set of criteria necessary for completing some task. Even complex tasks which take hundreds of steps like \texttt{Put all the mugs into the sink.} may be easily specified. When using the Planner's output, \model{} sequentially takes the next Planner action, runs the Object Detector and updates the knowledge representation, and replans. This allows the Planner to easily recover from incorrect initial detections or false negatives which may be corrected over time. The Planner returns control either when the goal state has been reached, when it determines the goal is impossible, or after a fixed number of steps. 

An example Planner sequence for the question \texttt{Is there a bowl in a drawer?} is as follows. The Meta-Controller invokes the Planner with the goal \texttt{All drawers have been checked or a bowl is in a drawer}. The knowledge state contains three drawers and the microwave. The Planner outputs a plan to check each drawer. The first drawer is empty, so the plan continues. In the second drawer, the bowl is found, and the Planner returns control to the Meta-Controller.

In a different example for the same task, the Planner checks each drawer and the bowl is in none of them. After all drawers are checked, the Planner returns control to the Meta-Controller.

Grouping the multi-step output from the Planner into a single decision made by the hierarchical controller gives \model{} several advantages over pure RL solutions. Primarily, this significantly reduces our action space and the number of high-level steps in a single episode. Furthermore, because the Planner guarantees that the goal will be reached (or ruled impossible), \model{} can be more thorough and efficient in its exploration. In the examples above, if the agent had only checked a single drawer and moved on, it would have missed the opportunity to know the answer was ``yes'' or be more confident that the answer was ``no.''

In this work we use the Metric-FF PDDL solver \cite{metricff}, one of the most popular planning algorithms for operating on PDDL instances. Metric-FF extends the origin FF Planning algorithm~\cite{ffplanner} to both boolean and numerical values. Metric-FF uses hill-climbing on relaxed plans (plans in which contradictions are ignored) as a heuristic to estimate the distance to the goal. For numeric values, the relaxation takes the form of ignoring any non-monotonicly increasing effects that an action may have. Finding the absolute shortest solution to PDDL problems is NP-Complete, but in practice, Metric-FF usually returns nearly optimal plans in around 100 milliseconds. We include a sample PDDL state and action domain and corresponding Planner output in the supplemental material.

\subsection{Object Detector}
\vspace{-2mm}
The Object Detector must detect objects from the current image, but it must also track what it has detected in the past. In this work, we assume perfect camera location knowledge which simplifies this process. The Object Detector predicts object masks as well as pixelwise depths, and the objects are projected into a global coordinate frame and merged with prior detections. In our experiments we use Mask-RCNN~\cite{maskrcnn} for the detection masks and the FRCN depth estimation network~\cite{laina2016deeper} which are both finetuned on the training environments. We merge detections by joining the bounding cube around two detections if their 3D intersection is above a certain threshold (in practice 0.3). A more sophisticated strategy with more frequent detections such as Fusion++\cite{mccormac2018fusion} could further improve our method (however Fusion++ requires a depth camera). In Figure~\ref{fig:network}, the Detected Object Map represents the previously detected objects and their spatial locations.

\subsection{Navigator}
\vspace{-2mm}
We use the Navigator from \cite{iqa} as it shows reliable performance for going to locations specified in a relative coordinate frame by a hierarchical controller. The Navigator predicts edge weights for a small region in front of the agent and uses an Egocentric Spatial GRU to update the memory state. Then it chooses the next action based on A* search to the target location. For more details, see \cite{iqa}. To improve the overall execution speed of our method, our algorithm only calls the Object Detector once the navigation has finished rather than at every intermediate location. 

\subsection{Stopper}
\vspace{-2mm}
\label{section:Stopper}
The Stopper is tasked with finishing an episode. For VSP, the Stopper simply terminates the episode. However for IQA, the Stopper must provide an answer to the posed question. For this, we train a network which takes the question, the entire memory state,  and the current image features as input and outputs an answer. For questions from IQUAD~V1, we use state information from the Object Detector to improve our accuracy. For example, for the question \texttt{Is there bread in the room?} since we track whether we have detected bread to be able to end planning upon detection, we can provide this information to the Stopper as well. For EQA V1, since the questions can be answered from a single image, we provided an additional image channel representing a detection mask of the question's subject. The Stopper is only trained based on the last state from a sequence via cross entropy over the possible answers. In practice, the Stopper shares a network with the Meta-Controller, as shown in Figure \ref{fig:network} which encourages the learned features to be semantically grounded.

\subsection{Explorer}
\vspace{-2mm}
To gather more information about an environment, the Explorer finds a location which has not been visited and invokes the Navigator with that location. In this work, the Explorer is not learned; instead, it tracks where the agent has been and picks the location and orientation which maximizes new views while minimizing the distance from the current agent location. Note that the Explorer still operates on the Navigator's learned free-space map.

\subsection{Scanner}
\vspace{-2mm}
The Scanner issues a predefined sequence of commands to the environment to obtain a 360\textdegree{} view of its surroundings. Specifically, it performs three full rotations at +30, 0, and -30 degrees with the horizon, stopping every 90 degrees to run the Object Detector. It is often useful to call the Scanner after calling the Explorer, but we leave this up to the hierarchical controller to learn.

%% file: 04_tasks.tex
\section{Tasks and Environments}
\label{section:tasks}
We focus on two tasks (Interactive Question Answering and Visual Semantic Planning) in two virtual environments (AI2-THOR~\cite{ai2thor} and House3D~\cite{house3d} for IQA, and AI2-THOR for VSP). Both tasks require complex visual and spatial reasoning as well as multi-step planning. 

\subsection{Interactive Question Answering (IQA)}
\vspace{-2mm}
We evaluate our agent on both IQUAD~V1~\cite{iqa} and EQA~V1~\cite{eqa}. IQUAD~V1 provides 75,600 training questions in 25 training rooms, and 1920 test questions in 5 unseen rooms. Additionally, IQUAD~V1 provides 2400 test questions in seen rooms which helps factor out errors due to object detection noise. For EQA~V1, we use the published train/val splits which consist of 7129 training questions in 648 train houses and 853 validation questions in 68 unseen houses. \cite{eqa, das2018neural} show results with the agent starting 10, 30, or 50 steps away from the goal, yet only their results for 10 steps outperform the language baselines of \cite{thomason2018shifting}. As such, we limit our experiments to starting 10 steps away. 

In IQUAD~V1, the agent must interact with kitchen scenes (opening drawers, exploring the room) in order to gather information to answer questions about the objects in the room such as \texttt{Is there a mug in the microwave?} EQA~V1 places an agent in a house a certain distance away from the object of interest and asks questions like \texttt{What color is the ottoman?} Both datasets use templated language. For more detail on these datasets, see \cite{eqa, iqa}.

For IQUAD~V1, we specify the goal state for the Planner as either checking all locations where the question's subject could be, or knowing where the question's subject is. For example, for the question \texttt{Is there a mug in the microwave?} the Planner must only check the microwave, but for the question \texttt{Is there a mug in the room?} the Planner must check the microwave, the fridge, and all the drawers until it finds at least one mug. For EQA~V1, we state the goal as ``the agent is looking at the object of interest if it knows where it is, or it has looked through all the doorways that it has seen.'' This encourages the Planner to explore various different rooms in the environment in the case where the object of interest is not located in the starting room.

\subsection{Visual Semantic Planning (VSP)}
\vspace{-2mm}
\label{section:vsp}
We also use the AI2-THOR environment~\cite{ai2thor} for semantic planning tasks such as \texttt{Put the apple in the sink.} These tasks are similar to those often done in Task and Motion Planning. Each task specifies one of 13 small objects (such as mug, fork, potato), and one of 6 target locations (such as fridge, cabinet, sink), but unlike in \cite{vsp}, we train one network to accomplish all tasks rather than one for each pair of object-locations. Additionally, we include navigation as a subtask of planning which \cite{vsp} omits. Compared with IQA, VSP contains only a single task type but uses a larger action space to facilitate picking up and putting down objects. Although we only test a single task type, we include more complex tasks in the supplemental material. We do not perform Visual Semantic Planning in House3D~\cite{house3d} as the environments are static. We use the same train/test split of environments as in IQUAD~V1, and construct 25,200 training tasks, 640 test tasks in unseen rooms, and 800 tasks in seen rooms. When constructing the tasks, we verify that each task is possible, yet cannot be completed by the empty plan, e.g. for the task \texttt{Put a mug in a cabinet.} there exists at least one mug and at least one cabinet, but no mugs start in cabinets. This data will be made available upon publication.

%% file: 05_experiments.tex
\section{Experiments}

\begin{table*}[t]
\begin{center}
\resizebox{\textwidth}{!}{
\begin{tabular}{lccc|ccc|ccc}
\hline
 & \multicolumn{3}{c|}{IQUAD V1} & \multicolumn{3}{c|}{EQA V1} & \multicolumn{3}{c}{VSP} \\
 & Accuracy & Episode Length & SSPL & Accuracy & Episode Length & SSPL & Success & Episode Length & SSPL \\ \hline
\begin{tabular}[c]{@{}l@{}}Planner With Global Information \\ (Shortest Path Estimate)\end{tabular} & 100\% & 88.710 & 1 & 100\% & 10 & 1 & 100\% & 87.477 & 1 \\ \hline
State-of-the-art for IQUAD~V1 \cite{iqa} and EQA~V1\cite{das2018neural} & 52.52\% & 586.890 & 0.015 & 53.58\% & - & - & - & - & - \\
Planner Only & 56.91\% & \textbf{138.644} & \textbf{0.105} & 49.53\% & \textbf{\phantom{0}44.424} & 0.002 & 11.41\% & \textbf{105.559} & 0.059 \\
\model{} & \textbf{65.99\%} & 357.690 & 0.086 & \textbf{58.41\%} & 154.781 & \textbf{0.007} & \textbf{46.01\%} & 427.784 & \textbf{0.189} \\ \hline
\cite{iqa} with GT Detections & 64.27\% & 531.840 & 0.042 & - & - & - & - & - & - \\
Planner Only with GT Detections & 74.53\% & \textbf{169.773} & \textbf{0.251} & 54.15\% & \textbf{\phantom{0}22.780} & \textbf{0.025} & 43.44\% & \textbf{161.998} & 0.245 \\
\model{} With GT Detections & \textbf{81.25\%} & 297.238 & 0.177 & \textbf{65.28\%} & 127.011 & 0.017 & \textbf{73.75\%} & 254.367 & \textbf{0.362} \\ \hline
\cite{iqa} with GT Detections and Nav & 69.85\% & 655.100 & 0.046 & - & - & - & - & - & - \\
Planner Only with GT Detections and Nav & 68.07\% & \textbf{\phantom{0}56.961} & 0.091 & 49.64\% & \textbf{\phantom{0}18.784} & 0.004 & 31.72\% & \textbf{\phantom{0}48.095} & 0.268 \\
\model{} with GT Detections and Nav & \textbf{83.30\%} & 182.191 & \textbf{0.325} & \textbf{65.52\%} & \phantom{0}65.237 & \textbf{0.033} & \textbf{81.88\%} & 161.013 & \textbf{0.549} \\ \hline
\end{tabular}
}
\end{center}
\vspace{-3mm}
\caption{Comparison of accuracy and episode length with varying levels of ground truth (GT) information on unseen environments. In some cases, the ``Planner Only'' is fast enough to outperform \model{} on the SSPL metric, indicating there is still significant progress to be made on speeding up \model{}. Interestingly, using the navigation system instead of GT navigation helps the ``Planner Only'' method by giving it more varied locations to run Object Detector. The shortest path estimate for IQUAD~V1 and VSP is equivalent to the ``Planner Only with GT Detections and Nav'' method except that it is additionally given the positions of all large objects (fridges, cabinets, etc.). In IQUAD~V1 finding true shortest paths is an instance of the traveling salesman problem. For EQA~V1 the shortest path is found using an oracle with preexisting knowledge of the location of the target object. In all experiments we use either the FRCN~\cite{laina2016deeper} depth network in conjunction with Mask-RCNN~\cite{maskrcnn} or ground truth detection masks and depth.}
\label{table:ablation}
\vspace{-3mm}
\end{table*}

We compare \model{} across the datasets outlined in Section \ref{section:tasks} using existing state-of-the-art methods as baselines as well as the unimodal baselines from \cite{thomason2018shifting} and pure planning baselines. On each dataset, we record the accuracy/success of our method as well as the episode length. Except in the generalization experiment, all tests are done on unseen environments. \cite{anderson2018evaluation} proposes the Success weighted by (normalized inverse) Path Length (SPL) which combines accuracy and episode length into a single metric for evaluating embodied agents on pure navigation tasks. SPL is defined as 
\begin{equation}
    SPL = \frac{1}{N}\sum_{i=1}^N{S_i \frac{\ell_i}{max(p_i, \ell_i)}}
\end{equation}
\noindent where $S_i$ is a success indicator for episode $i$, $p_i$ is the path length, and $\ell_i$ is the shortest path length. SPL is not sufficient for question answering as an agent which never moves could still be very successful depending on the difficulty of the questions\footnote{\cite{thomason2018shifting} shows significant bias exists in EQA~V1~\cite{eqa} and in MatterPort3D~\cite{anderson:cvpr18}}. To address this issue, we propose the Shifted SPL (SSPL) metric which is defined as
\begin{equation}
    SSPL = \frac{\mu - b}{1 - b} * SPL
\end{equation}

\noindent where $\mu$ is the average accuracy of the method and $b$ is the average accuracy of a baseline agent which is forced to end/answer immediately after beginning an episode. Note that SSPL directly accounts for dataset biases by subtracting the accuracy of a learned baseline rather than simply the most common answer or random chance accuracy. For the VSP experiments SPL is exactly equal to SSPL, as a baseline which cannot move will achieve 0\% success. For the IQA experiments we use the ``Language Only'' baselines presented in \cite{thomason2018shifting} as $b$.

\subsection{Baselines}
\vspace{-2mm}
On all datasets we include (at least) one pure-learning and one pure-planning baseline. The ``Planner Only'' baseline uses the same Plan/Act/Observe/Replan loop as \model{} but does not include any hierarchical decision making. Additionally, if at the start of the episode the plan is empty (for example if the agent starts looking at a wall), we rotate the agent until the plan is not empty. We also use the ``Language Only'' baselines from \cite{thomason2018shifting} which attempt to answer the question without making any actions, effectively learning the language bias of the dataset. For VSP, we introduce a ``Learning Only'' baseline which removes the Planner from \model{} and adds reward shaping to encourage certain interactions like looking at and picking up the object of the task. Even after significant training time, this method fails to learn a working policy.

\begin{figure}[t]
\begin{center}
\includegraphics[width=1.0\linewidth]{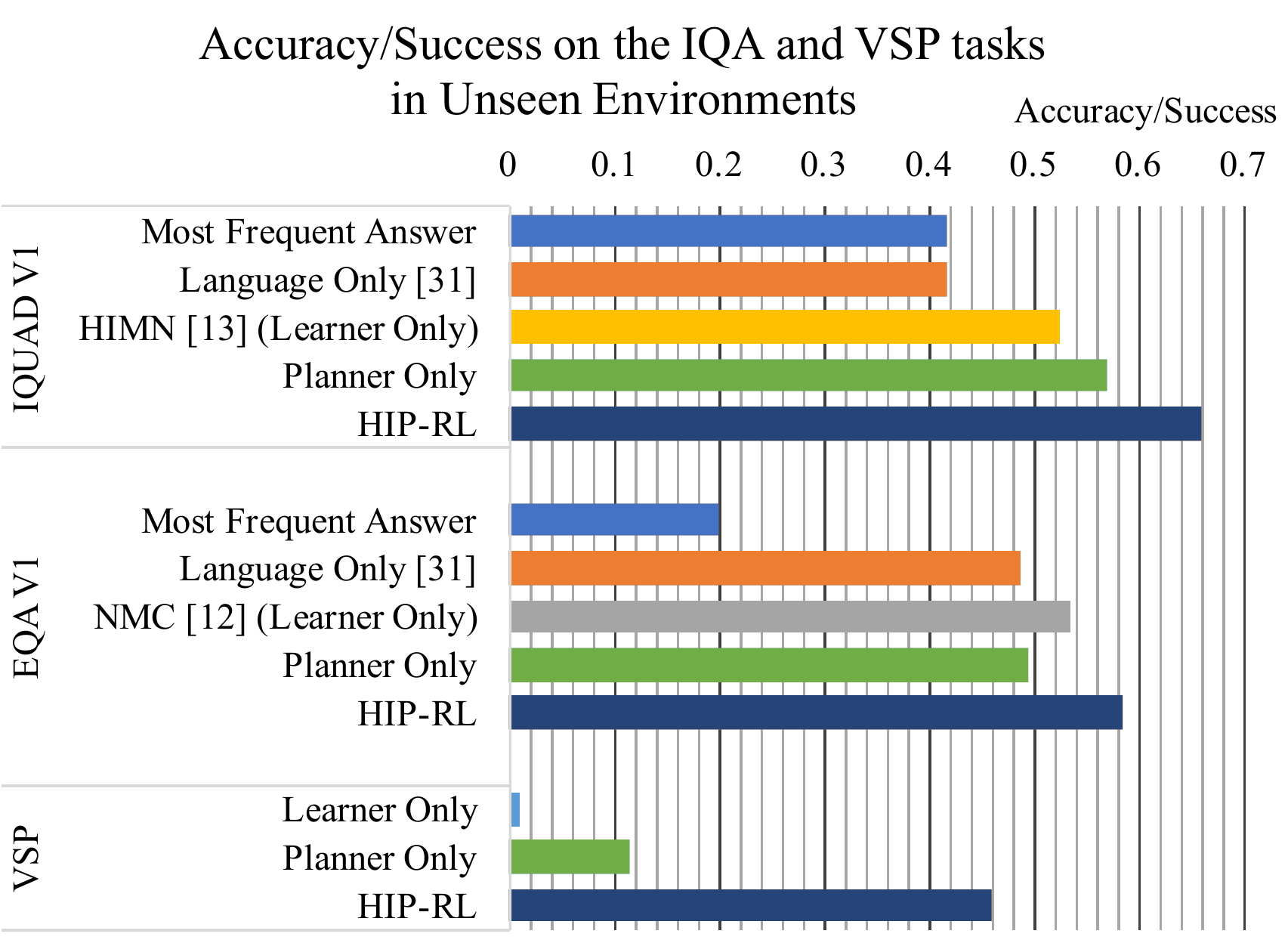}
\vspace{-7mm}
\caption{Accuracy of various methods on each of the tasks. In all cases, \model{} achieves state-of-the-art performance. We include ``Learner Only'' and ``Planner Only'' results for each experiment to show that combining both their strengths is better than either alone.} 
\label{fig:bar_chart}
\end{center}
\vspace{-7mm}
\end{figure}

\subsection{Results}
\vspace{-2mm}
We test our system for accuracy on IQUAD~V1, EQA~V1, and the new VSP dataset, achieving state-of-the-art performance on all tasks. The results are shown in Figure \ref{fig:bar_chart}. In IQUAD~V1 and VSP the ``Planner Only'' baseline outperforms the ``Learning Only'' baseline which coincides with the fact that the ground-truth trajectories are significantly longer and contain more necessary interactions than in EQA~V1. A fundamental issue with reinforcement learning is that it must ``luck into'' good solutions randomly before it can improve, which can be very rare in complex multi-step tasks. Planning simplifies this by directly solving objectives rather than making guesses and observing rewards or penalties. Yet pure planning suffers from myopia in that (in our system) it assumes perfect and complete global information, leading it to ignore unobserved parts of the environment. This is most apparent in VSP where the planner assumes a task is impossible if it has not observed a location where the object can be. By combining both strategies, \model{} achieves the exploration tendencies of RL along with the goal-oriented direct problem solving of planning.

\subsection{Ablation}
\vspace{-2mm}
We further explore the effect that various sources of inaccuracy have on \model{} by substituting the Object Detector and the Navigator with Ground Truth (GT) information, shown in Table \ref{table:ablation}. Adding GT detections greatly improves our accuracy across the board. This is due to all the tasks being very object-centric, so if the object is misidentified or not detected at all, the Answerer/Planner has no means of fixing the mistake. In contrast, using GT Navigation does not improve performance dramatically, but the path lengths do nearly halve. In practice we observe this is frequently not from the navigation agent wandering randomly but is instead usually from the beginning of the episodes where the map starts empty and the navigator unknowingly goes down dead ends or takes otherwise inefficient paths.

\subsection{Episode Efficiency}
\vspace{-2mm}
Table \ref{table:ablation} also lists episode lengths and SSPL scores for each method. Note that episode lengths include every interaction with the environment (turn left, open, move ahead each count as one action), not just hierarchical actions. While \model{} dramatically improves over~\cite{iqa}, there is still a large gap between the shortest path estimate. Some inefficiency due to exploration is unavoidable, but there are also cases where the agent explores even after it could answer. This generally occurs in IQUAD~V1 on counting questions where the agent is not sure that it has sufficiently checked everywhere where the object could be.

\begin{table}[t]
\begin{center}
\resizebox{\columnwidth}{!}{
\begin{tabular}{l|ccc|ccc}
\hline
 & \multicolumn{3}{c|}{IQUAD V1 Unseen} & \multicolumn{3}{c}{IQUAD V1 Seen} \\
 & Accuracy & Length & SSPL & Accuracy & Length & SSPL \\
\model{} & 65.99\% & 357.690 & 0.086 & 77.75\% & 265.668 & 0.182 \\
\model{} + GT Det & 81.25\% & 297.238 & 0.177 & 87.04\% & 277.538 & 0.278 \\ \hline \hline
 & \multicolumn{3}{c|}{VSP Unseen} & \multicolumn{3}{c}{VSP Seen} \\
 & Success & Length & SSPL & Success & Length & SSPL \\
\model{} & 46.01\% & 427.784 & 0.189 & 70.88\% & 245.301 & 0.384 \\
\model{} + GT Det & 73.75\% & 254.367 & 0.362 & 82.48\% & 170.22 & 0.504 \\ \hline
\end{tabular}
}
\end{center}
\vspace{-3mm}
\caption{Comparison of accuracy/success on seen and unseen environments. \model{} is the full method, and \model{} + GT Det uses the ground truth detections.}
\label{table:generalization}
\vspace{-3mm}
\end{table}

\subsection{Generalization}
\vspace{-2mm}
One benefit of hierarchical models is they tend to generalize better as they force certain structure to be consistent between seen and unseen environments. Yet if the hierarchical models depend on the performance of individual components, then the generalization performance of the constituent models directly affects the overall performance. In Table \ref{table:generalization} we explore the generalization of \model{} on IQUAD~V1 and VSP by comparing performance on rooms seen during training time with never-before-seen rooms (EQA~V1 only provides test questions for unseen environments). With ground truth detections, \model{} generalizes quite well, losing less than 10\% raw performance in both cases and staying nearly as efficient step-wise. With Mask-RCNN~\cite{maskrcnn} detections and FRCN depth~\cite{laina2016deeper}, performance is still reasonably similar, but there is a larger gap. Mask-RCNN produces high-quality results on large datasets, yet in the case of AI2-THOR~\cite{ai2thor}, there are frequently only 25 training and 5 testing samples of a particular class. Thus, Mask-RCNN struggles to detect the cabinets and drawers in unseen environments from AI2-THOR (as there are many cabinets and drawers per scene but none repeat in multiple rooms), so frequently many areas remain unchecked. We believe that in scenarios with many more training examples, \model{} with detection would approach the same level of generalization performance as without.

\subsection{Learning Speed}
\vspace{-2mm}
We compare the convergence speed of \model{} on the IQUAD~V1 task with the previous state-of-the-art model, HIMN~\cite{iqa}. After only 26,000 hierarchical steps, \model{} with ground truth information matches the final performance of HIMN with ground truth at 8 million hierarchical steps\footnote{
We use the number of hierarchical steps rather than the total number of steps in the environments as hierarchical steps are of variable length and do not provide gradients to the hierarchical controller until the final low-level action.}. After 120,000 hierarchical steps, \model{} (without ground truth) converges to its maximum performance compared to HIMN which takes 500,000 iterations and achieves significantly worse performance. \model{} trains orders of magnitude faster than traditional RL algorithms because the planner simplifies much of the learning by being able to immediately (upon initialization) return good, thorough trajectories. Being thorough early on is especially useful for question answering where an algorithm may get confusing feedback if it answers too soon; for example, for the question \texttt{Is there a bowl in the room?} if an agent does not see a bowl and answers ``no'' but the correct answer is ``yes,'' the network will receive contradictory learning signals. By using a planner, we ensure more thorough exploration so this case is much less likely to occur, even at the beginning of training.

\begin{figure}[t]
\begin{center}
\includegraphics[width=1.0\linewidth]{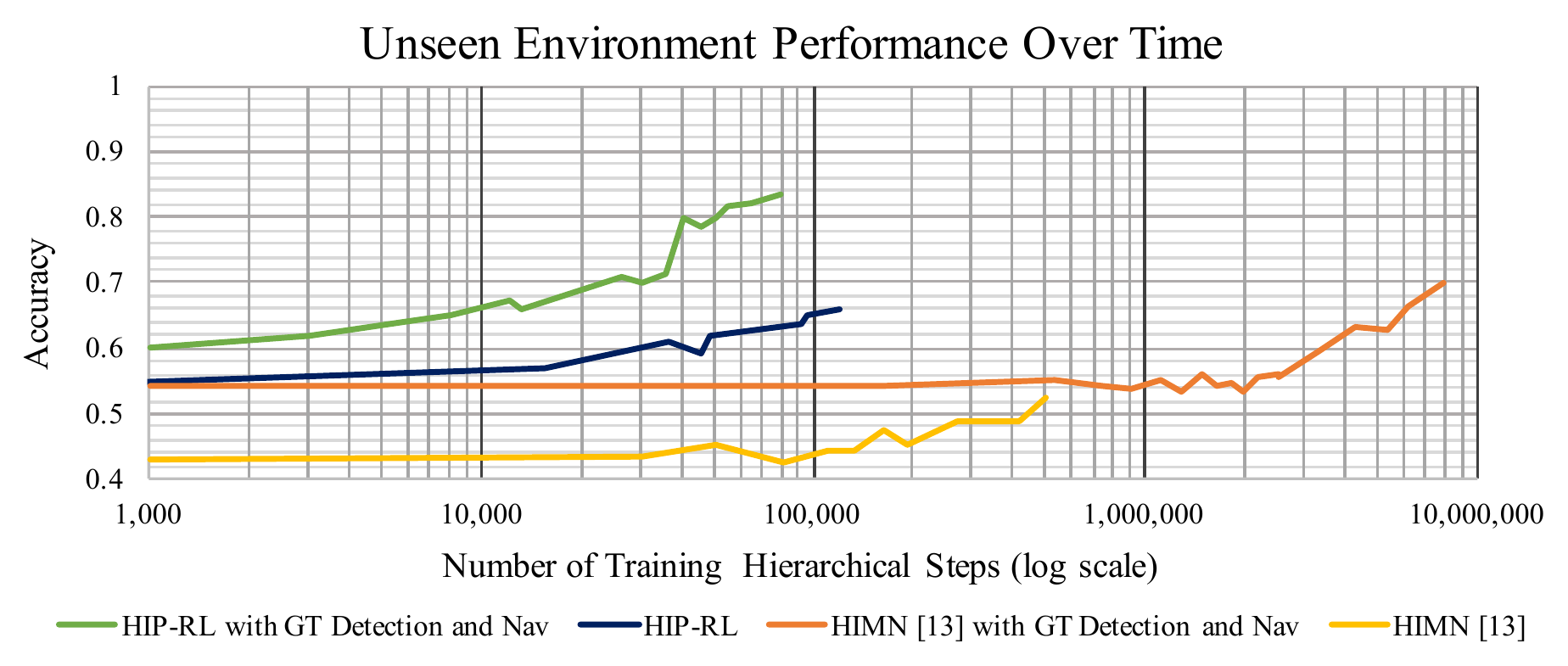}
\vspace{-7mm}
\caption{Learning speed of \model{} and HIMN~\cite{iqa} with and without ground truth information. HIMN's Answerer is additionally pretrained on fully observed rooms whereas \model{} does not require any pretraining.} 
\label{fig:learning_speed}
\end{center}
\vspace{-8mm}
\end{figure}

%% file: 06_conclusion.tex
\section{Conclusion}
In this work, we presented \modellong{}, a method for combining the benefits of Deep Reinforcement Learning and Symbolic Planning. We demonstrate its effectiveness at increasing accuracy while simultaneously decreasing episode length and training time. Though this exact implementation may not be applicable to many other tasks, we believe the high level idea of learning to invoke various direct controllers, some of which explicitly plan, could be applied to a broader array of tasks such as Task and Motion Planning. In general, we observe that using planning algorithms to assist in performing ``good'' actions results in improved accuracy, test-time efficiency, and training speed. Still, \model{} could be improved to explore more efficiently using priors based on likely locations of an object. Additionally, we could learn the PDDL preconditions and effects directly so as to limit the need for human labels. We are excited about the potential impacts of visual agents and their ability to learn to interact more intelligently with the world around them.

\section{Acknowledgements}
This work was funded in part by the National Science Foundation under contract number NSF-NRI-1637479, NSF-IIS-1338054, NSF-1652052, ONR N00014-13-1-0720, the Allen Distinguished Investigator Award, and the Allen Institute for Artificial Intelligence. We would like to thank NVIDIA for generously providing a DGX used for this research via the UW NVIDIA AI Lab (NVAIL). 

%% file: 07_appendix.tex
\appendix
\clearpage
\newpage
\begin{appendices}
\section{PDDL Domain}
\noindent Below is the full PDDL Domain for question answering and visual semantic planning. 

{\fontsize{6}{7.2}\selectfont
\begin{lstlisting}
(define (domain qa_vsp_task)
  (:requirements
    :adl
  )
  (:types
    agent
    location
    receptacle
    object
    rtype
    otype
  )
  
  (:predicates
    (atLocation ?a - agent ?l - location)
    (receptacleAtLocation ?r - receptacle ?l - location)
    (objectAtLocation ?o - object ?l - location)
    (openable ?r - receptacle)
    (opened ?r - receptacle)
    (inReceptacle ?o - object ?r - receptacle)
    (checked ?r - receptacle)
    (receptacleType ?r - receptacle ?t - rtype)
    (objectType ?o - object ?t - otype)
    (canContain ?t - rtype ?o - otype)
    (holds ?a - agent ?o - object)
    (holdsAny ?a - agent)
    (full ?r - receptacle)
  )

  (:functions
    (distance ?from ?to)
    (totalCost)
  )

  ;; agent goes to receptacle
  (:action GotoLocation
    :parameters (?a - agent ?lStart - location ?lEnd - location)
    :precondition (atLocation ?a ?lStart)
    :effect (and
      (atLocation ?a ?lEnd)
      (not (atLocation ?a ?lStart))
      (forall (?r - receptacle)
        (when (and (receptacleAtLocation ?r ?lEnd) 
                   (or (not (openable ?r)) (opened ?r)))
          (checked ?r)
        )
      )
      (increase (totalCost) (distance ?lStart ?lEnd))
    )
  )

  ;; agent opens receptacle
  (:action OpenObject
    :parameters (?a - agent ?l - location ?r - receptacle)
    :precondition (and
      (atLocation ?a ?l)
      (receptacleAtLocation ?r ?l)
      (openable ?r)
      (forall (?re - receptacle)
        (not (opened ?re)))
    )
    :effect (and
      (opened ?r)
      (checked ?r)
      (increase (totalCost) 1)
    )
  )
  
  ;; agent closes receptacle
  (:action CloseObject
    :parameters (?a - agent ?l - location ?r - receptacle)
    :precondition (and
      (atLocation ?a ?l)
      (receptacleAtLocation ?r ?l)
      (openable ?r)
      (opened ?r)
    )
    :effect (and
      (not (opened ?r))
      (increase (totalCost) 1)
    )
  )

  ;; agent picks up object
  (:action PickupObject
    :parameters (?a - agent ?l - location ?o - object ?r - receptacle)
    :precondition (and
      (atLocation ?a ?l)
      (objectAtLocation ?o ?l)
      (or (not (openable ?r)) (opened ?r))
      (inReceptacle ?o ?r)
      (not (holdsAny ?a))
    )
    :effect (and
      (not (inReceptacle ?o ?r))
      (holds ?a ?o)
      (holdsAny ?a)
      (increase (totalCost) 1)
    )
  )

  ;; agent puts down object
  (:action PutObject
    :parameters (?a - agent ?l - location ?ot - otype ?o - object ?r - receptacle)
    :precondition (and
      (atLocation ?a ?l)
      (receptacleAtLocation ?r ?l)
      (or (not (openable ?r)) (opened ?r))
      (not (full ?r))
      (objectType ?o ?ot)
      (holds ?a ?o)
    )
    :effect (and
      (inReceptacle ?o ?r)
      (full ?r)
      (not (holds ?a ?o))
      (not (holdsAny ?a))
      (increase (totalCost) 1)
    )
  )
)

\end{lstlisting}
}
\section{PDDL Goal Example}
\noindent Below is the goal specification for the question \texttt{Is there a mug in the room?}. 

{\fontsize{6}{7.2}\selectfont
\begin{lstlisting}
(:goal
  (or
    (exists (?o - object)
      (objectType ?o MugType))
    (and
      (forall (?t - rtype) 
        (forall (?r - receptacle)
          (or
            (not (and (canContain ?t MugType)
                      (receptacleType ?r ?t)))
            (checked ?r)
          )
        )
      )
      (forall (?re - receptacle)
        (not (opened ?re)))
    )
  )
)
\end{lstlisting}
}
\end{appendices}